\documentclass[conference]{IEEEtran}
\IEEEoverridecommandlockouts
\usepackage{amsmath,amssymb,amsfonts}
\usepackage{algorithmic}
\usepackage{graphicx}
\usepackage{textcomp}
\usepackage{xcolor}
\usepackage[numbers]{natbib}
\usepackage{amsthm}
\usepackage{booktabs}
\usepackage{subcaption}
\usepackage[numbers]{natbib}
\usepackage{tikz}
\usepackage{float}

\theoremstyle{definition}

\def\BibTeX{{\rm B\kern-.05em{\sc i\kern-.025em b}\kern-.08em
    T\kern-.1667em\lower.7ex\hbox{E}\kern-.125emX}}
    
\makeatletter
\newcommand*\titleheader[1]{\gdef\@titleheader{#1}}
\AtBeginDocument{%
  \let\st@red@title\@title
  \def\@title{%
    \bgroup\normalfont\large\centering\@titleheader\par\egroup
    \vskip1.5em\st@red@title}
}

\title{Not All Adversarial Examples Require a Complex Defense: Identifying Over-optimized Adversarial Examples with IQR-based Logit Thresholding}
\titleheader{\textregistered \, IEEE 2019 \\ Accepted for the 2019 International Joint Conference on Neural Networks}

\author{
\IEEEauthorblockN{Utku Ozbulak$^{1,\,3}$ \quad Arnout Van Messem$^{2,\,3}$ \quad Wesley De Neve$^{1,\,3}$}
\IEEEauthorblockA{\textit{$^1$Department of Electronics and Information Systems, Ghent University, Belgium} \\
\textit{$^2$Department of Applied Mathematics, Computer Science and Statistics, Ghent University, Belgium} \\
\textit{$^3$Center for Biotech Data Science, Ghent University Global Campus, Republic of Korea}\\
\{utku.ozbulak,arnout.vanmessem,wesley.deneve\}@ugent.be}

}

\begin{document}


\maketitle

\begin{abstract}
Detecting adversarial examples currently stands as one of the biggest challenges in the field of deep learning. Adversarial attacks, which produce adversarial examples, increase the prediction likelihood of a target class for a particular data point. During this process, the adversarial example can be further optimized, even when it has already been wrongly classified with $100\%$ confidence, thus making the adversarial example even more difficult to detect. For this kind of adversarial examples, which we refer to as \textit{over-optimized} adversarial examples, we discovered that the logits of the model provide solid clues on whether the data point at hand is adversarial or genuine. In this context, we first discuss the masking effect of the softmax function for the prediction made and explain why the logits of the model are more useful in detecting over-optimized adversarial examples. To identify this type of adversarial examples in practice, we propose a non-parametric and computationally efficient method which relies on interquartile range, with this method becoming more effective as the image resolution increases. We support our observations throughout the paper with detailed experiments for different datasets (MNIST, CIFAR-10, and ImageNet) and several architectures.
\end{abstract}




\section{Introduction}
Even though deep convolutional architectures outperform other models on various image-related problems such as image classification \cite{resnet}, object detection \cite{DBLP:journals/corr/RedmonDGF15}, and segmentation \cite{DBLP:journals/corr/RonnebergerFB15}, it has been shown that these models are not foolproof. A recent development called \textit{adversarial examples} currently counts as one of the major issues these models are facing. Adversarial examples were first discovered by \cite{Szegedy-Intriguingproperties} while searching for intriguing properties of neural networks. Although there is no clear definition of \textit{adversarial example}, we will say that a data point is called adversarial if it is perturbed to be misclassified.

The earliest methods to generate adversarial examples mostly rely on optimizing an input for a specific target class using a gradient-based algorithm, hence forcing the data point under consideration to be classified in another category \cite{Szegedy-Intriguingproperties, Nguyen-deepnnseasilyfooled}. To improve the effectiveness of adversarial example generation, a variety of methods based on different optimization techniques were introduced. Some examples are: finding the closest distance to the decision boundary \cite{moosavi2016deepfool}, Fast-Gradient Sign (FGS)~\cite{Goodfellow-expharnessing}, Jacobian-based Saliency Maps (JSMA)~\cite{DBLP:journals/corr/PapernotMJFCS15}, and the recently introduced \textit{Carlini \& Wagner attack} (CW)~\cite{DBLP:journals/corr/CarliniW16a}. The latter is currently regarded as the strongest attack, given that it can incorporate defense mechanisms into the optimization procedure, thus making it possible to generate strong adversarial examples that are able to circumvent these defense mechanisms~\cite{DBLP:journals/corr/CarliniW16a}.

As attacks get more effective with every study, new defense mechanisms are also introduced to counter them. The intuitive approach of adversarial re-training was the first method tested to prevent this problem \cite{Goodfellow-expharnessing}. \cite{DBLP:journals/corr/KurakinGB16a} applied adversarial re-training on ImageNet \cite{ILSVRC15:rus} and found that it counters certain attacks. Another mechanism to prevent adversarial examples was introduced by \cite{DBLP:journals/corr/PapernotMWJS15}, showing that network distillation provides some defense against adversarial examples generated with JSMA. \cite{DBLP:journals/corr/LuIF17} suggested extracting a binary threshold from the output of each rectifier layer (ReLU) and utilizing these thresholds with a quantized radial basis support vector machine detector to distinguish adversarial examples from genuine images. The most impactful research on adversarial defense mechanisms has thus far been performed by \cite{DBLP:journals/corr/CarliniW17}, who showed that none of the proposed defense mechanisms so far are effective against strong attacks and that the properties of currently available defense mechanisms do not scale well to higher resolution images, e.g., from MNIST images to ImageNet images.


\cite{DBLP:journals/corr/CarliniW17} further argued that defense mechanisms should be evaluated against strong attacks and that they need to demonstrate that white-box attacks can be prevented. In this context, an attack is said to be white-box in nature if the attacker has access to the specifications of the model (weights and classes) and is aware of the defense mechanism. However, assuming the attacker has access to both the model and the defense mechanism presents an unfair challenge to the defender. Indeed, the knowledge about the model and the defense mechanism can then be incorporated into an attack, making it possible to generate adversarial examples that have been carefully designed to avoid the defense mechanism under consideration. In their evaluation of defense techniques, \cite{DBLP:journals/corr/CarliniW17} put two constraints on the generation of adversarial examples, namely (1) discretization and (2) the amount of perturbation. We believe this approach may lead to results that need to be carefully interpreted because, as we will show in this paper, it is trivial to distinguish natural images from heavily optimized adversarial examples based on logit values. Therefore, we claim that not all adversarial examples necessarily require a complex defense mechanism and that, in fact, a significant number of them can already be detected by analyzing the logits of the prediction.

\textbf{Contributions:} In summary, our work makes the following contributions:
\begin{itemize}
\item We discuss the masking effect of the softmax function, hiding the magnitude of the predicted logit value. This observation lays the foundation for a novel technique for identifying adversarial examples, as discussed in more detail in the remainder of this paper.
\item We show that attack mechanisms may generate what we call \textit{over-optimized} adversarial examples. These adversarial examples, which have extremely high logit outputs, can be easily identified when the logits of the prediction are analyzed.
\item We introduce a non-parametric method to calculate a logit threshold from training data. This threshold can then be used to identify over-optimized adversarial examples.
\item When the resolution of an image increases, we show that the subspace in which adversarial examples can be generated also increases substantially. Furthermore, unlike some of the proposed defense techniques, our method is able to identify more and more adversarial examples as the resolution of the given image increases. We provide evidence for these observations by making use of the MNIST, CIFAR-10, and ImageNet datasets.
\end{itemize}


\section{Impact of Adversarial Optimization on Logits}
Methods for adversarial example generation use specific optimization techniques to increase the chance a given input is labeled with the targeted class. Most of the proposed optimization techniques are iterative in nature. Indeed, it has been shown that these techniques produce more effective adversarial examples with less perturbation, compared to single-step optimization techniques or techniques that facilitate untargeted attacks~\cite{athalye2018obfuscated}.

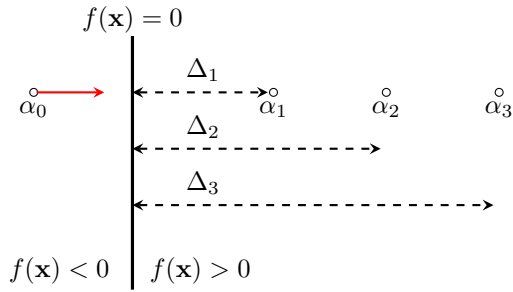
\begin{figure}
\begin{tikzpicture}[scale=0.75]
\draw[very thick] (4.5, 4.5) -- (4.5, 0);

\draw (4.5,  5.2) node[anchor=north]{$f(\mathbf{x})=0$};
\draw (5.7,  0.75) node[anchor=north]{$f(\mathbf{x})>0$};
\draw (3.2,  0.75) node[anchor=north]{$f(\mathbf{x})<0$};

\draw (2.75, 3.5) node[anchor=north]{$\mathbf{\alpha}_0$};
\draw [circle] (2.75, 3.5) circle[radius=2pt];
\draw[thick, ->,>=stealth,red] (2.8, 3.5) -- (4, 3.5);
\draw (5.75, 4.25) node[anchor=north]{$\Delta_1$};
\draw (7, 3.5) node[anchor=north]{$\mathbf{\alpha}_1$};
\draw [circle] (7, 3.5) circle[radius=2pt];
\draw[thick, dashed, stealth-stealth] (4.5, 3.5) -- (6.9, 3.5);

\draw (5.75, 3.25) node[anchor=north]{$\Delta_2$};
\draw (9, 3.5) node[anchor=north]{$\mathbf{\alpha}_2$};
\draw [circle] (9, 3.5) circle[radius=2pt];
\draw[thick, dashed, stealth-stealth] (4.5, 2.5) -- (8.9, 2.5);
\draw (5.75, 2.25) node[anchor=north]{$\Delta_3$};
\draw (11, 3.5) node[anchor=north]{$\mathbf{\alpha}_3$};
\draw [circle] (11, 3.5) circle[radius=2pt];
\draw[thick, dashed, stealth-stealth] (4.5, 1.5) -- (10.9, 1.5);

\end{tikzpicture}
\caption{An illustration of adversarial examples $\alpha_{\{1,2,3\}}$, derived from an input $\alpha_0$, for an affine classifier $f(x)=0$,  and their respective distances $\Delta_{\{1,2,3\}} = d(\alpha_{\{1,2,3\}}, f)$ to the decision boundary~$f$.}
\vspace{-1em}
\label{linear_classifier}
\end{figure}

We can simplify the behavior of targeted iterative optimization techniques as follows: assume that we have an affine classifier in a two-dimensional setting, $f(\mathbf{x}):\mathbb{R}^2\to\mathbb{R}$, separating the input space into two subspaces $f(\mathbf{x})>0$ and $f(\mathbf{x})<0$, with the line $f(\mathbf{x}) = 0$ denoting the decision boundary. This setting is shown in Fig.~\ref{linear_classifier}. Under these circumstances, when the given example $\alpha_0$ is iteratively optimized towards $f(x)>0$, the distances $\Delta_{\{1,2,3\}}$ between the generated adversarial points $\alpha_{\{1,2,3\}}$ and the decision boundary are increased at each step, with the goal to increase the likelihood that the data point under consideration (that is, $\alpha_0$) is classified as $f(\textbf{x})>0$. To that end, it does not matter whether the attack uses the logits of the model (like CW) or not, given that the overall impact of the optimization techniques used on the logits remains the same. A more detailed explanation of adversarial optimization and its impact on the decision boundary for multi-class classifiers can be found in~\cite{moosavi2016deepfool}.

In this study, we consider two targeted attacks: (1) Basic Iterative Method~\cite{DBLP:journals/corr/KurakinGB16}, a fast attack that arguably generates \textit{weak} adversarial examples, and (2) CW, a slow attack that generates \textit{strong} adversarial examples.

\subsection{Basic Iterative Method}
Basic Iterative Method, also referred to as Iterative Fast Gradient Sign (I-FGS), finds its origin in the FGS method proposed in \cite{Goodfellow-expharnessing}. FGS performs a one-step update on the input along the direction of the gradient:
\begin{align*}
\mathbf{x}_{n+1} =  \mathbf{x}_{n} - \alpha \ \text{sign} (\nabla_x J(g(\theta,\mathbf{x}_n)_c) \,,
\end{align*}
where $\text{sign} (\nabla_x J(.))$ is the signature of the gradient of the cross-entropy error function and $\alpha$ is the perturbation multiplier. This method is characterized as \textit{fast} since it does not require an iterative approach. 

\cite{DBLP:journals/corr/KurakinGB16} extended FGS into I-FGS, proposing to use a lower value for $\alpha$ and to iteratively update the input image:
\begin{align*}
\mathbf{x}_{n+1} =  Clip_{\mathbf{X}, \epsilon}(  \mathbf{x}_{n} - \alpha \ \text{sign} (\nabla_x J(g(\theta,\mathbf{x}_n)_c)  \,.
\end{align*}
In the above equation, the clipping function makes sure that the resulting adversarial example is a valid image. Furthermore, the parameter $\epsilon$ controls the maximally allowed amount of perturbation per pixel.

FGS, along with its extension I-FGS, is one of the most commonly used adversarial attacks to test adversarial defense mechanisms. However, in this study, we opted for I-FGS over FGS because I-FGS is shown to produce stronger adversarial examples~\cite{DBLP:journals/corr/KurakinGB16}.


  \begin{figure*}[t]
\captionsetup[subfigure]{justification=centering}
    \centering
      \begin{subfigure}{0.22\textwidth}
        \includegraphics[width=\textwidth]{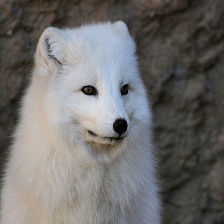}
          \caption{Original Image \\ Prediction: Arctic Fox \\Confidence: $0.99$ \\ $\text{Logit}_1$: $\sim 20$ \\ $\text{Logit}_2$: $\sim 5 \phantom{0}$}
      \end{subfigure}
      \begin{subfigure}{0.22\textwidth}
        \includegraphics[width=\textwidth]{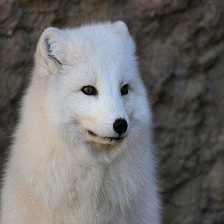}
          \caption{Adv. Image\\ Prediction: Radio \\Confidence: $1$ \\ $\text{Logit}_1$: $\sim 1e2$ \\ $\text{Logit}_2$: $\sim 10 \phantom{0}$}
      \end{subfigure}
      \begin{subfigure}{0.22\textwidth}
        \includegraphics[width=\textwidth]{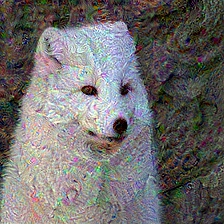}
          \caption{Adv. Image\\  Prediction: Radio \\Confidence: $1$ \\ $\text{Logit}_1$: $\sim 5e2$ \\ $\text{Logit}_2$: $\sim 12 \phantom{0}$}
      \end{subfigure}
      \begin{subfigure}{0.22\textwidth}
        \includegraphics[width=\textwidth]{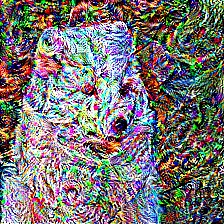}
          \caption{Adv. Image\\  Prediction: Radio \\Confidence: $1$ \\ $\text{Logit}_1$: $\sim 4e3$ \\ $\text{Logit}_2$: $\sim 40 \phantom{0}$}
      \end{subfigure}
      \caption{(a) Original image, predicted as \textit{arctic fox} with $0.99$ confidence. (b)-(c)-(d) Over-optimized adversarial examples which are predicted with the same confidence but with vastly different logit values by ResNet-50. $\text{Logit}_1$ and $\text{Logit}_2$ represent the logits of the most likely and second most likely predictions, respectively.}
      \label{fig:over-opt}
      \vspace{-1em}
\end{figure*}



\subsection{Carlini \& Wagner Attack}
Even though FGS and I-FGS can generate adversarial examples with a high success rate, the main focus of these methods is not to produce \textit{strong} adversarial examples. In~\cite{DBLP:journals/corr/CarliniW16a}, Carlini and Wagner proposed a heavily optimized attack which produces strong adversarial examples that can bypass defense mechanisms easily. In this study, we use the $L_2$ version of CW, which was also used by Carlini and Wagner to test the robustness of defense techniques \cite{DBLP:journals/corr/CarliniW17}. The $L_2$ version of CW is defined as follows:
\begin{align*}
\text{miminize} \quad & ||\mathbf{x} - (\mathbf{x} + \delta)||_{2}^{2} + \alpha  \; \ell(\mathbf{x} + \delta)\,,
\end{align*}
where this attack attempts to find a small perturbation $\delta$ that is sufficient to change the prediction made by the model when it is added to the input, while keeping the $L_2$ distance between the original image $\mathbf{x}$ and the perturbed image $\mathbf{x}' = \mathbf{x} + \delta$ minimal.

In their original work \cite{DBLP:journals/corr/CarliniW16a}, they discussed multiple loss functions for this kind of attack, and in this study, we use the loss function they preferred in their later work \citep{DBLP:journals/corr/CarliniW17} to evaluate multiple defense mechanisms. This loss function $\ell$ is constructed as follows:
\begin{align}
\label{eq:CW2} 
\ell (\mathbf{x}') = \max \left\{  \max \{ g(\theta, \mathbf{x}')_i : i\neq c\} - g(\theta, \mathbf{x}')_c, -  \mathit{K} \right\} \,,
\end{align}
comparing the logit prediction of the target class with the logit prediction of the next-most-likely class. This method can also be tuned using $\mathit{K}$ to adjust the strength of the adversarial example produced. Specifically, as $\mathit{K}$ increases, the confidence of the prediction for the adversarial example also increases. They refer to this kind of adversarial examples as \textit{high-confidence adversarial examples}.

\section{Masking Effect of the Softmax Function}
When the prediction of a neural network is analyzed, the output is usually represented in terms of probabilities. Such probability is usually referred to as the confidence of the prediction made. To convert logit values into probabilities, a normalized exponential function called the softmax function $P(\mathbf{u})_k = \dfrac{e^{\mathbf{u}_k}}{\sum_{m=1}^{M}e^{\mathbf{u}_m}}$ is used, where $\mathbf{u}$ is an input vector such that $\mathbf{u}=(u_1, \ldots, u_M)^T \in \mathbb{R}^{M}$ and $k$ is the selected index of the vector $\mathbf{u}$~\citep{Bishop06a,Goodfellow-et-al-2016,bridle1990probabilistic}. In particular, the softmax function uses the exponential function to squeeze the input values between zero and one in such a way that the output values add up to one. This property makes the softmax function helpful in more easily interpreting the predictions of a neural network, instead of having to rely on the logits, which are more difficult to interpret. 

The usage of the softmax function for convolutional neural networks dates back to 1998~\cite{lecun1998gradient}, soon thereafter becoming a common tool to convert logits into probabilitistic values. However, when investigating adversarial examples, the softmax function, when used in settings with limited decimal precision, has a major drawback for correctly interpreting the predictions of a neural network: it lacks sensitivity to positive changes in the magnitude of the largest logit. We detailed this discovery and its impact on adversarial examples in a previous study~\cite{ozbulak2018softmax}.



\subsection{Experimental Results on the Masking Effect of Softmax}
 
  \begin{figure*}[t]
\captionsetup[subfigure]{justification=centering}
    \centering
      \begin{subfigure}{0.45\textwidth}
        \includegraphics[width=\textwidth]{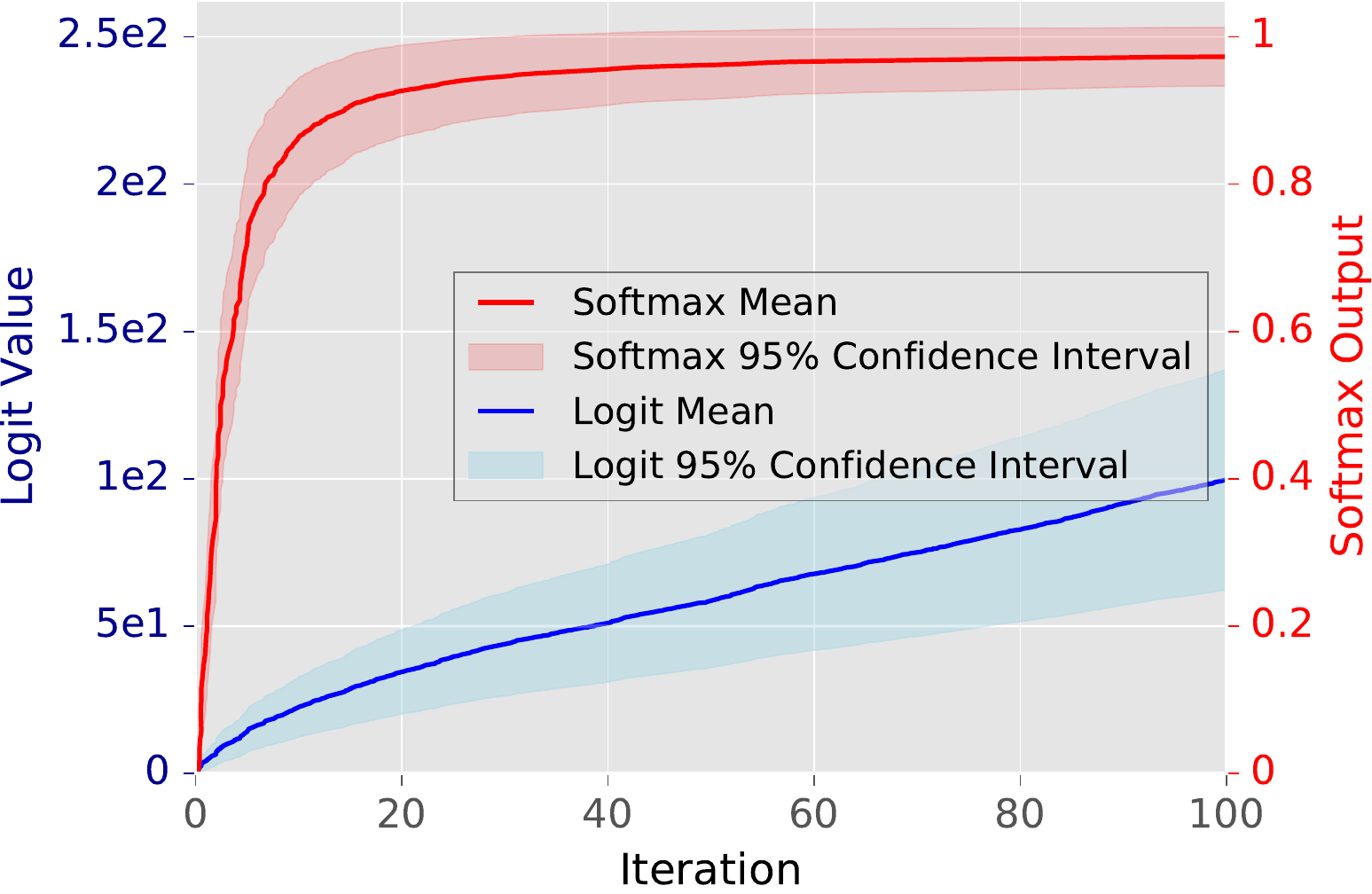}
          \caption{I-FGS}
      \end{subfigure}
      \begin{subfigure}{0.45\textwidth}
        \includegraphics[width=\textwidth]{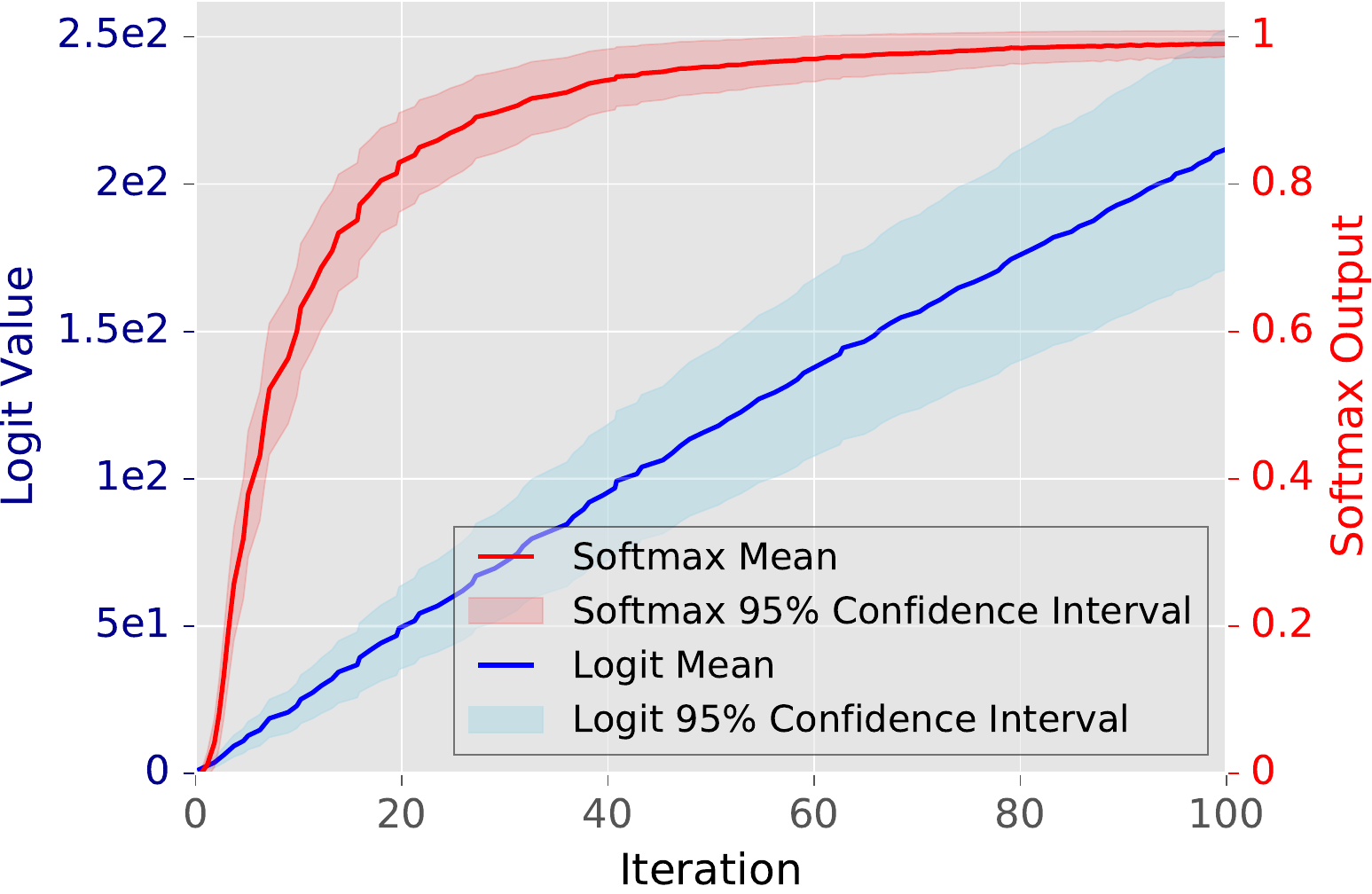}
          \caption{CW}
      \end{subfigure}
      \caption{Highest logit value and corresponding softmax output as a function of the number of iterations when generating adversarial examples with (a) I-FGS and (b) CW. Adversarial examples are tested on ResNet-50 in a white-box setting. Best viewed in color.}
      \label{fig:IFGS-CW}
      \vspace{-1em}
\end{figure*}

\begin{figure}[t]
  \includegraphics[width=0.48\textwidth]{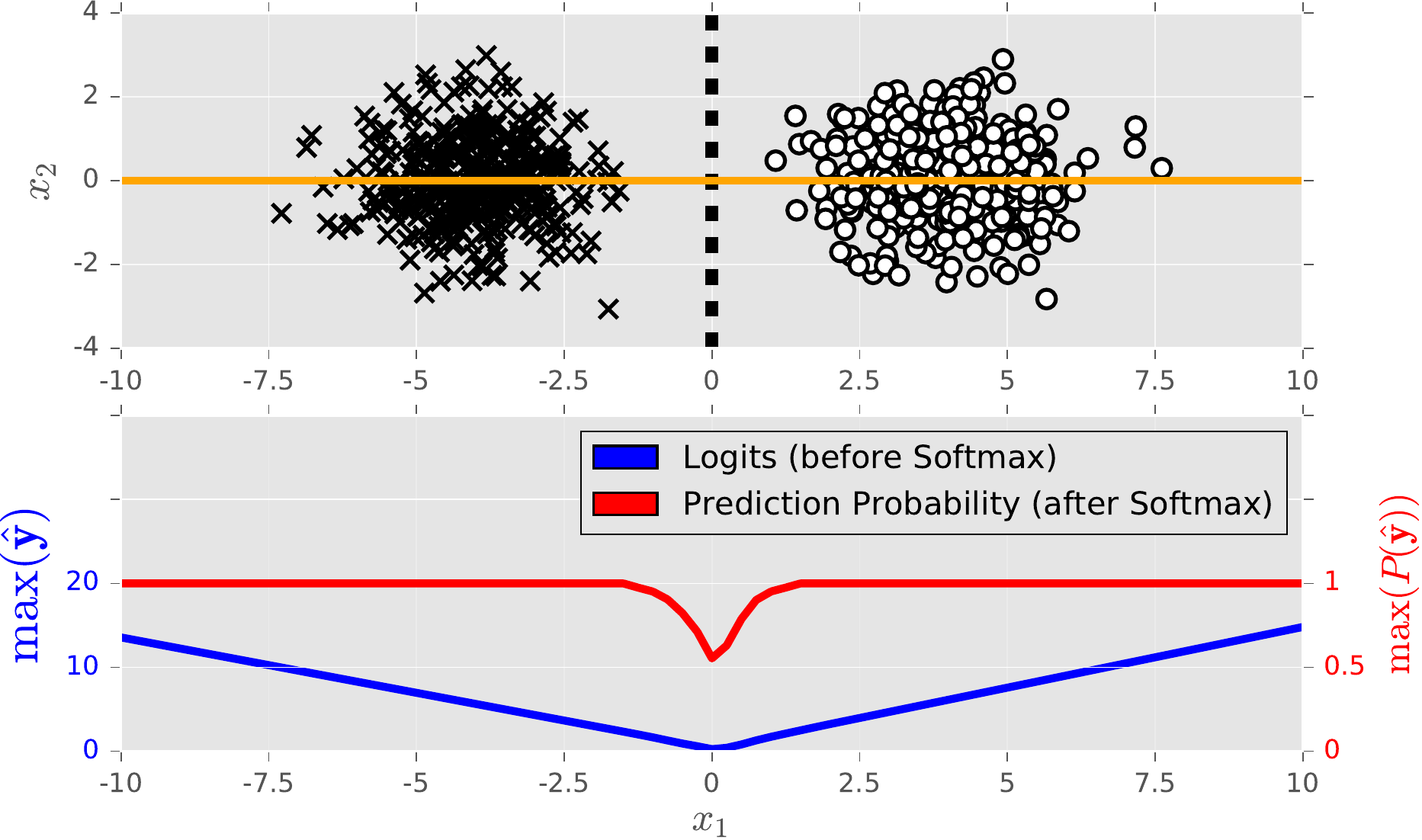}
          \caption{A linearly separable two-class classification problem, highlighting the saturation of the softmax output, as opposed to the logit values, which keep increasing. Best viewed in color.}
          \label{fig:ca-soft-line-chart}
          \vspace{-1em}
 \end{figure}

To show a practical outcome of the aforementioned limitation of the softmax function for neural networks, we present Fig.~\ref{fig:ca-soft-line-chart}, showing a mock-up two-class classification problem that visualizes the masking effect of the softmax function. A neural network $\hat{\mathbf{y}}=g(\theta, \mathbf{x})$ with a single hidden layer and ReLU activations is applied to this problem. The approximate decision boundary is indicated using a dashed line. For those points lying on the orange line $(x_2=0)$, Fig.~\ref{fig:ca-soft-line-chart} displays both the predicted logit value $\max(g(\theta, \mathbf{x}))$ and the predicted outcome probability $\max(P(g(\theta, \mathbf{x})))$. It can be seen from Fig.~\ref{fig:ca-soft-line-chart} that the logit values increase as the distance between the point under consideration and the classification boundary increases. However, the softmax output, after a certain point, remains the same and bounded. 


Another demonstration of the aforementioned limitation of the softmax function for natural images can be observed in Fig.~\ref{fig:over-opt}. In this figure, we provide an original image, classified as \textit{arctic fox} with a confidence value of $0.99$, and three adversarial counterparts, all of which are classified as \textit{radio} with a confidence value of $1$, and where classification is done by a pretrained ResNet-50 \cite{resnet}. All three adversarial examples have been over-optimized in order to produce increasing logits at each step. However, the softmax output for all of them remains the same, masking the increase in the logit values.


In Fig.~\ref{fig:IFGS-CW}, we present observations for multiple adversarial examples, showing the mean and the confidence interval of the highest logit value of $1000$ adversarial examples and their corresponding softmax output. The $1000$ adversarial examples were generated by making use of I-FGS and the $L_2$ version of the CW, using $100$ optimization steps in a white-box setting. For I-FGS, we perturb the image one pixel at a time, and for CW, we use $K=40$ in Eq.~\ref{eq:CW2}. The increase in the logit values for I-FGS is less pronounced than for CW since I-FGS relies on the signature of the gradient, which removes the precision in the perturbation between pixels. On the other hand, the spike in the confidence for I-FGS appears faster than for CW because the latter comes with a multi-class optimization nature and implements an extensive search process. Fig.~\ref{fig:IFGS-CW} clearly shows the masking effect of softmax, given that the confidence almost immediately jumps to $100\%$ after only a couple of iterations, making it from this point onwards impossible to differentiate between consecutive adversarial examples, whereas logit values keep increasing over the course of the optimization.

The results provided in Fig.~\ref{fig:IFGS-CW} can be generalized to other targeted iterative adversarial example generation methods \cite{ Nguyen-deepnnseasilyfooled, DBLP:journals/corr/CarliniW16a, Szegedy-Intriguingproperties, DBLP:journals/corr/PapernotMJFCS15, DBLP:journals/corr/KurakinGB16}. These iterative methods make it possible to further optimize an adversarial example (in terms of the logits), even after obtaining full confidence. However, once the prediction confidence is mapped to one, it is impossible to differentiate between the next iterations of the adversarial example based on the softmax output. Indeed, as the corresponding softmax input (i.e., logit values) keeps increasing, the softmax output will remain the same.

We refer the reader to our previous work which details the masking effect of the softmax function in the case of both single-target and multi-target adversarial attacks~\cite{ozbulak2018softmax}.

\label{ClassActivation_Softmax}
  \begin{figure*}[t]
\captionsetup[subfigure]{justification=centering}
    \centering
      \begin{subfigure}{0.24\textwidth}
        \includegraphics[width=\textwidth]{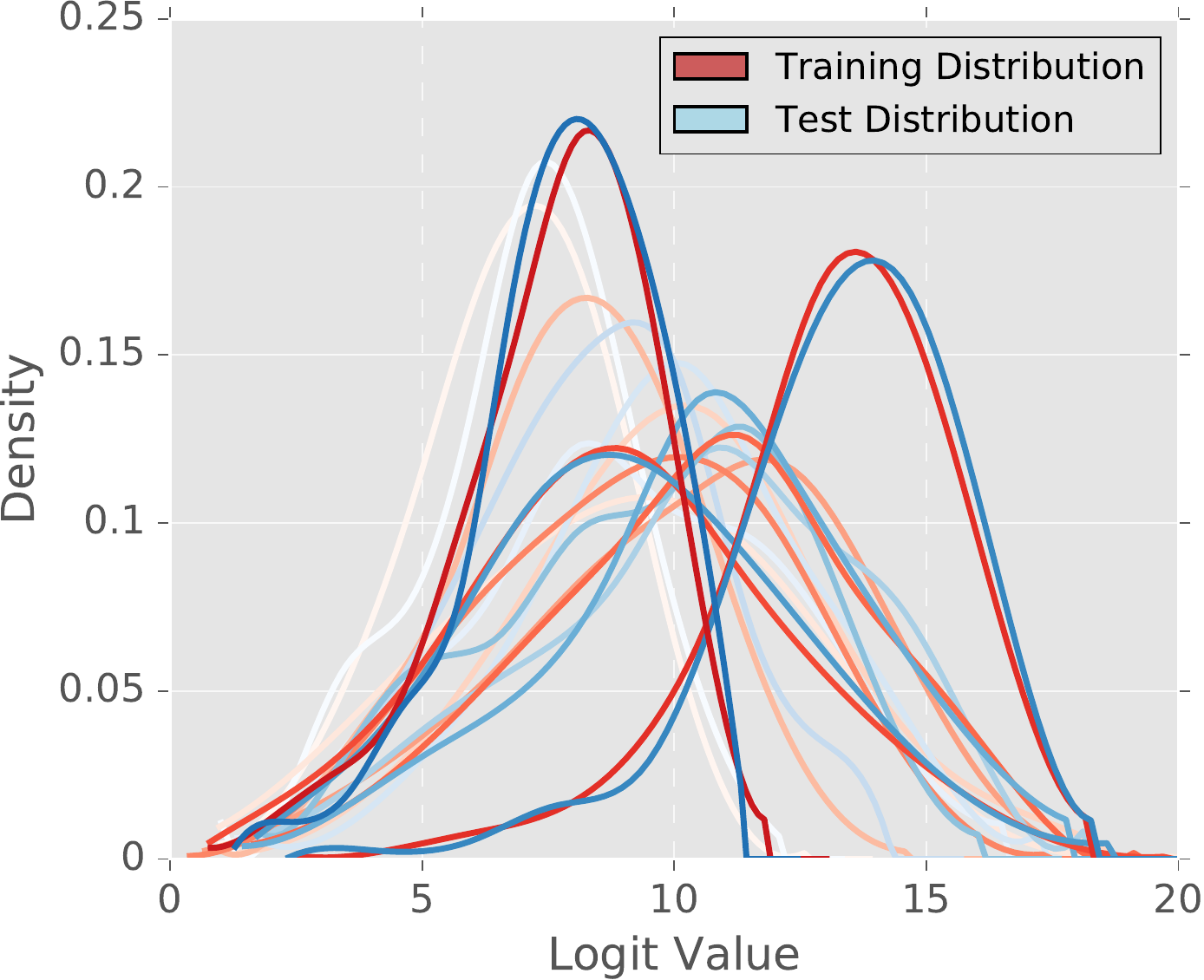}
        \caption{Dataset: MNIST\\Model: LeNet-5}
      \end{subfigure}
      \begin{subfigure}{0.24\textwidth}
        \includegraphics[width=\textwidth]{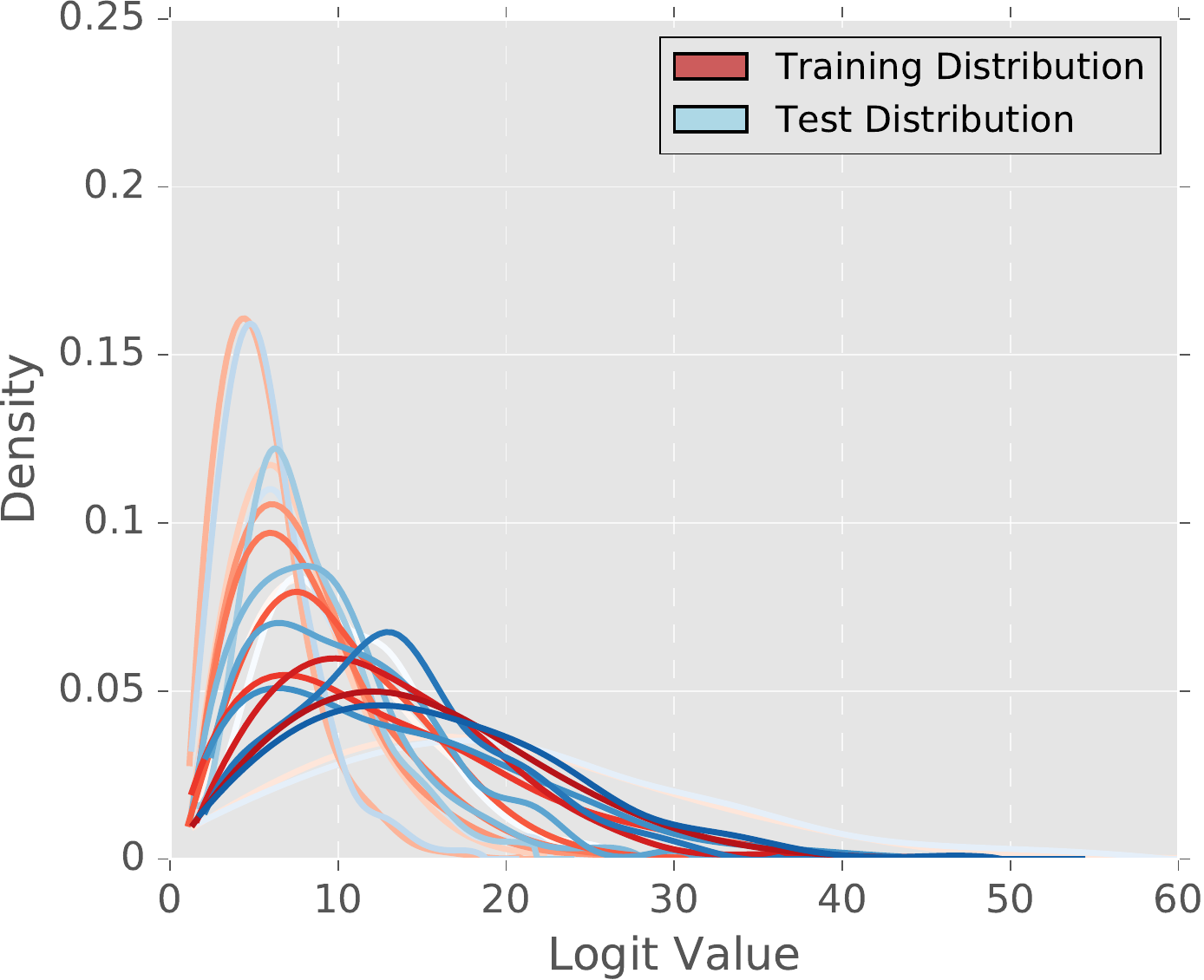}
        \caption{Dataset: CIFAR-10\\Model: Extended LeNet-5}
      \end{subfigure}
         \begin{subfigure}{0.24\textwidth}
        \includegraphics[width=\textwidth]{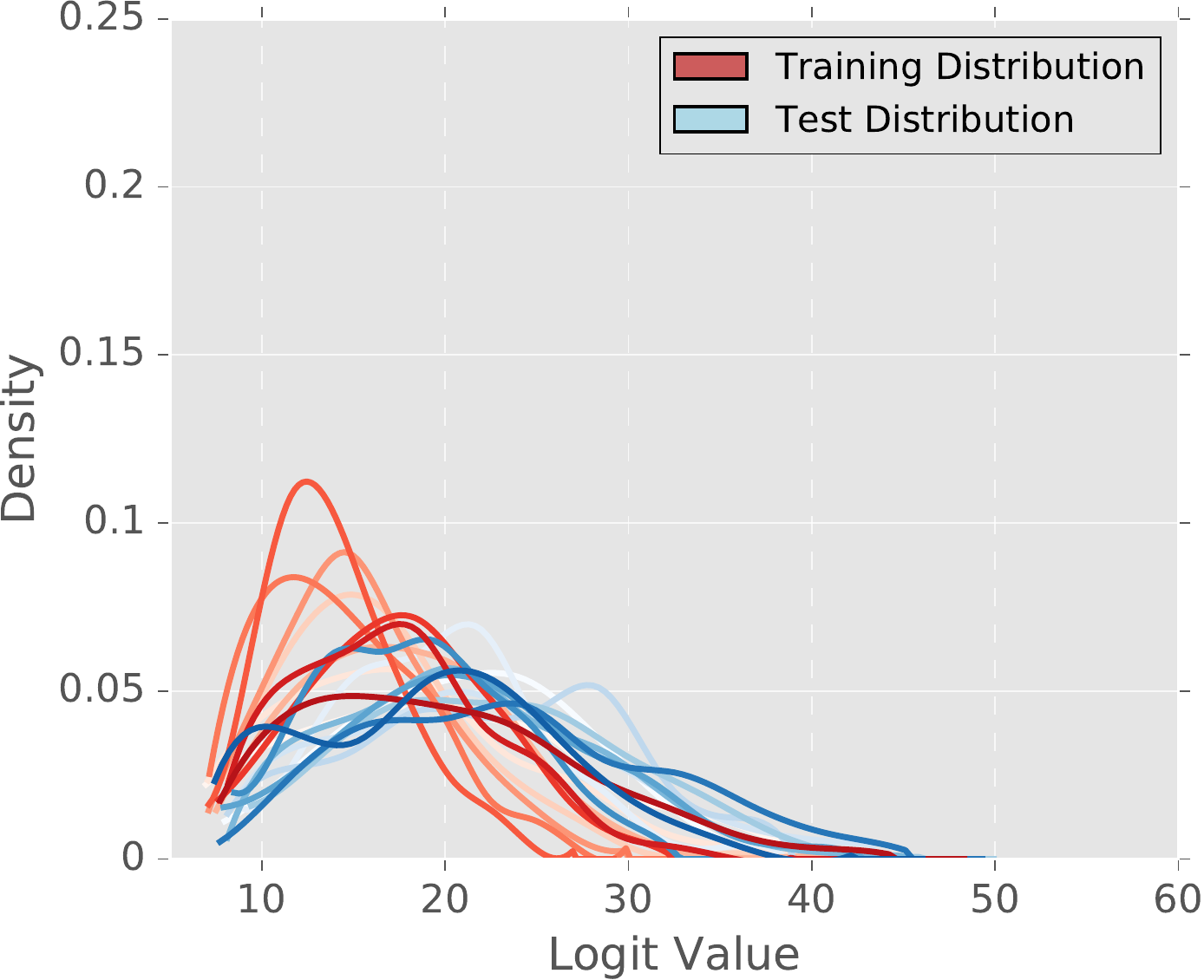}
        \caption{Dataset: ImageNet\\ Model: VGG-16}
      \end{subfigure}
         \begin{subfigure}{0.24\textwidth}
        \includegraphics[width=\textwidth]{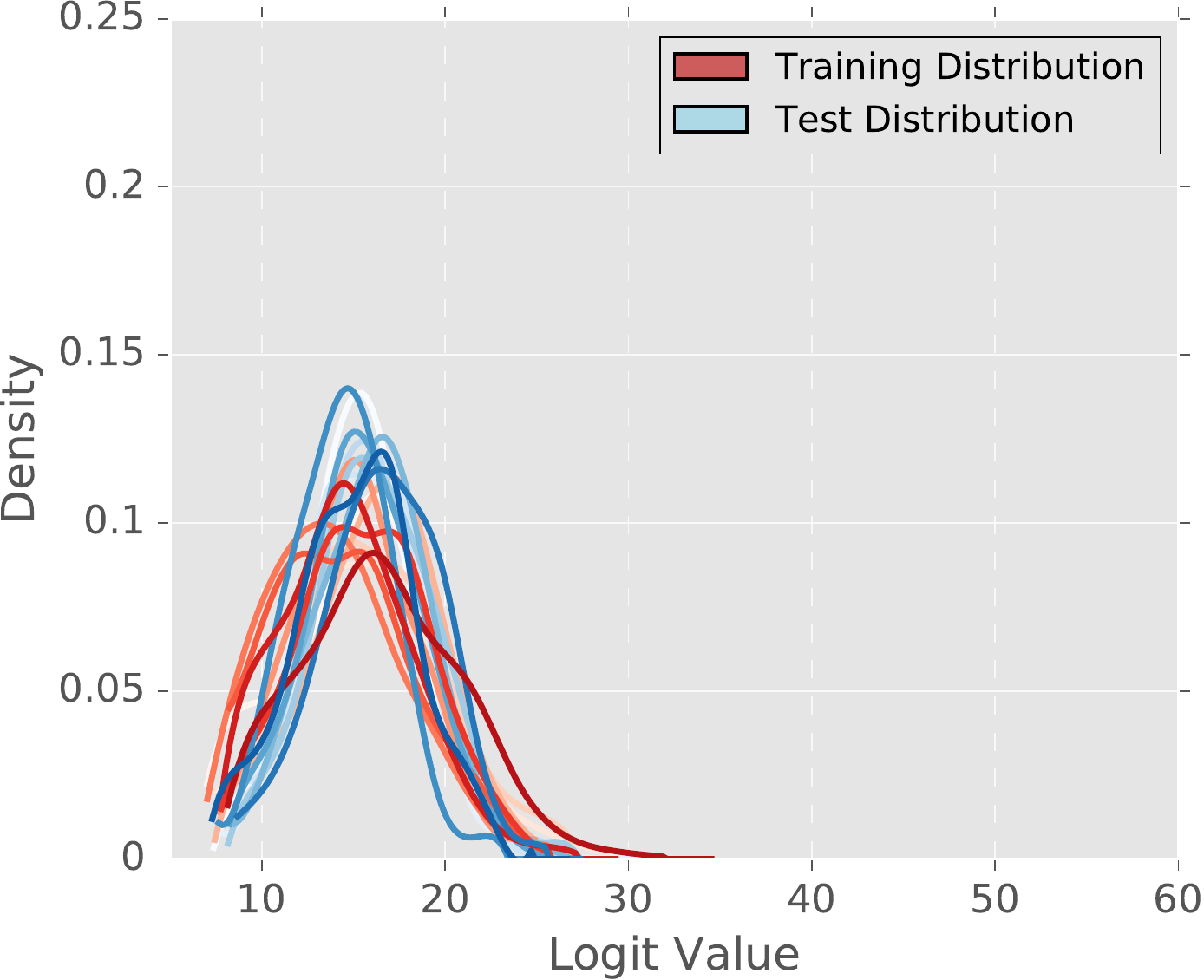}
        \caption{Dataset: ImageNet\\ Model: ResNet-50}
      \end{subfigure}
      \caption{Density plots of logit values of the prediction for ten classes, observed for both seen and unseen examples of the (a) MNIST, (b) CIFAR-10, and (c), (d) ImageNet datasets. We used LeNet-5 with ReLU activation~\citep{lecun1998gradient}, Extended LeNet-5 \cite{DBLP:journals/corr/PapernotMWJS15, DBLP:journals/corr/CarliniW17}, VGG-16 \cite{VGG}, and ResNet-50 \cite{resnet} to obtain these results. For ImageNet, ten classes were selected randomly. Our models achieved $98\%$, $80\%$, $70.5\%$, and $77\%$ top-1 accuracy on the respective datasets. These results are comparable to the results presented in \citep{resnet, DBLP:journals/corr/PapernotMWJS15, DBLP:journals/corr/CarliniW17, VGG}.}
      \label{fig:distributions}
\end{figure*}


\begin{table*}[t!]
  \centering
  \begin{tabular}{lcccccc}
    \toprule
    Dataset (Model)     & $1.5  \ IQR$  &  $2 \ IQR$ & $3 \ IQR$ & $4 \ IQR$ & $5 \ IQR$ & $k_{\min}$\\
    \midrule
    MNIST (LeNet-5) & $0.1\%$  & $0.004\%$   & $\mathbf{0}\%$ &$\mathbf{0}\%$ & $\mathbf{0}\%$  & $2.2 $\\
    CIFAR-10  (Extended LeNet-5)   & $1.7\%$  & $0.6\%$ & $0.07\%$   & $0.01\%$ &  $\mathbf{0}\%$ & $4.9$ \\
    ImageNet (VGG-16)    & $1.1\%$  & $0.3\%$  & $0.03\%$ & $0.004\%$ & $\mathbf{0}\%$ & $4.6$ \\
     ImageNet (ResNet-50)    & $0.7\%$  & $0.1\%$  & $0.009\%$ & $\mathbf{0}\%$ & $\mathbf{0}\%$  & $3.9$ \\
      ImageNet (Inception-v3)    & $1.2\%$  & $0.4\%$  & $0.07\%$ & $0.001\%$ & $\mathbf{0}\%$ & $4.9$  \\
    \bottomrule
  \end{tabular}
  \vspace{0.3em}
    \caption{Percentage of genuine images incorrectly identified as outliers, as obtained for different values of $k$ in the following calculation: $g(\theta , \mathbf{x}) \overset{?}{>} Q_3 + k \ IQR$. Thresholds are calculated from correctly classified training examples for each class and tested on both training and test examples. $k_{min}$ is the smallest value of $k$ needed to ensure that none of the examples in the training and the test set are incorrectly identified as outliers.}
\label{t:k-table}
\vspace{-1em}
\end{table*}

\section{Identifying Adversarial Examples}


Targeted attacks, in every form, focus on maximizing an activation; for neural networks, this activation is the logit value for a particular class. Unless this optimization is stuck in a local maximum, we can say that $g(\theta, \mathbf{x}_{i+1})_c \geq g(\theta , \mathbf{x}_{i})_c$, where $c$ is the targeted class. If $\mathbf{x}$ is a data point that is the subject of box constraints, then this activation can only be maximized up to a certain point. Let $\mathbf{x}_{r}$ be a hypothetical genuine image that achieves the highest logit value for its category in supervised settings. We are interested in finding the numerical value of $g(\theta , \mathbf{x}_{r})_c$, as this will allow us to label any data point that produces a higher logit value than this value as an adversarial example without any further evaluation.  To that end, in order to come up with an effective method to estimate $g(\theta ,\mathbf{x}_{r})_c$, we first analyze the prediction distributions of genuine images in terms of logit values.

\subsection{Logit Distributions}

In order to come up with a robust method for countering adversarial examples (that is, a method that works for all models across all datasets), we first analyze the logit distribution of genuine images. A reoccurring problem in the field of adversarial examples is the usage of low-resolution images to show the effectiveness of a defense mechanism. Indeed, not all of the defense techniques for low-resolution images transfer to high-resolution images \cite{DBLP:journals/corr/CarliniW17}. For this reason, we investigate the effectiveness of our method for MNIST, as well as for CIFAR-10 and ImageNet. In this context, we present Fig.~\ref{fig:distributions}, which shows the densities of predicted logit values for correctly classified samples of ten classes taken from MNIST, CIFAR-10, and ImageNet, for both training and test datasets. In the case of ImageNet, we present the distribution for both VGG-16 and ResNet-50. We can observe that the distributions of the logit values change between different datasets and between different classes/labels. On top of that, the distributions are vastly different for different architectures (VGG-16 and ResNet-50), even when the same dataset is used.

Keeping the aforementioned observations in mind, we arrive at the following conclusions: (a) the proposed method should be distribution-free so that it can be generalized across multiple datasets and multiple models, and (b) since the distribution of logit values also changes between different classes, the computational complexity of the proposed method should be low. 

  \begin{figure*}[t]
\captionsetup[subfigure]{justification=centering}
    \centering
      \begin{subfigure}{0.49\textwidth}
        \includegraphics[width=\textwidth]{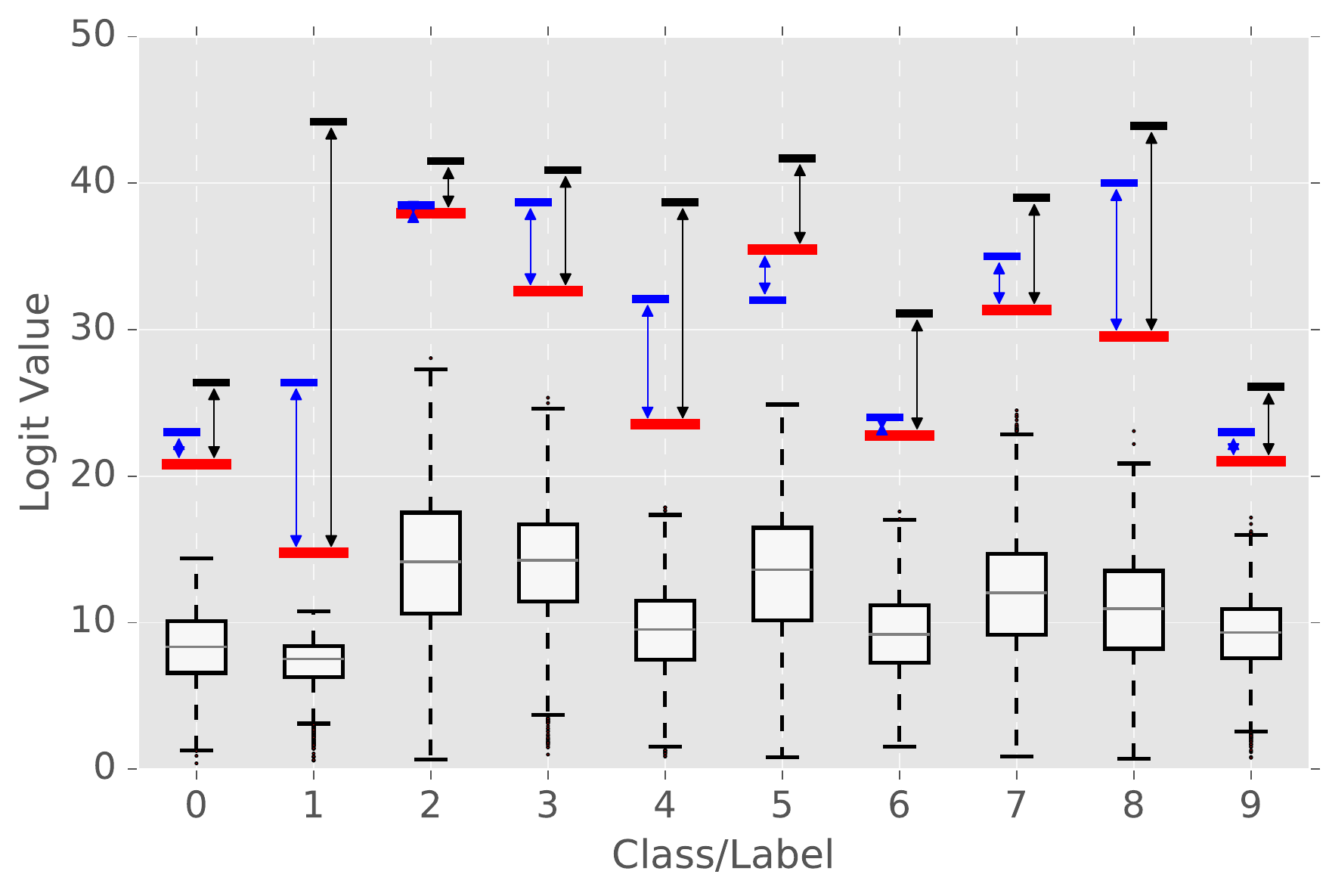}
          \caption{Dataset: MNIST\\Model: LeNet-5}
      \end{subfigure}
      \begin{subfigure}{0.49\textwidth}
        \includegraphics[width=\textwidth]{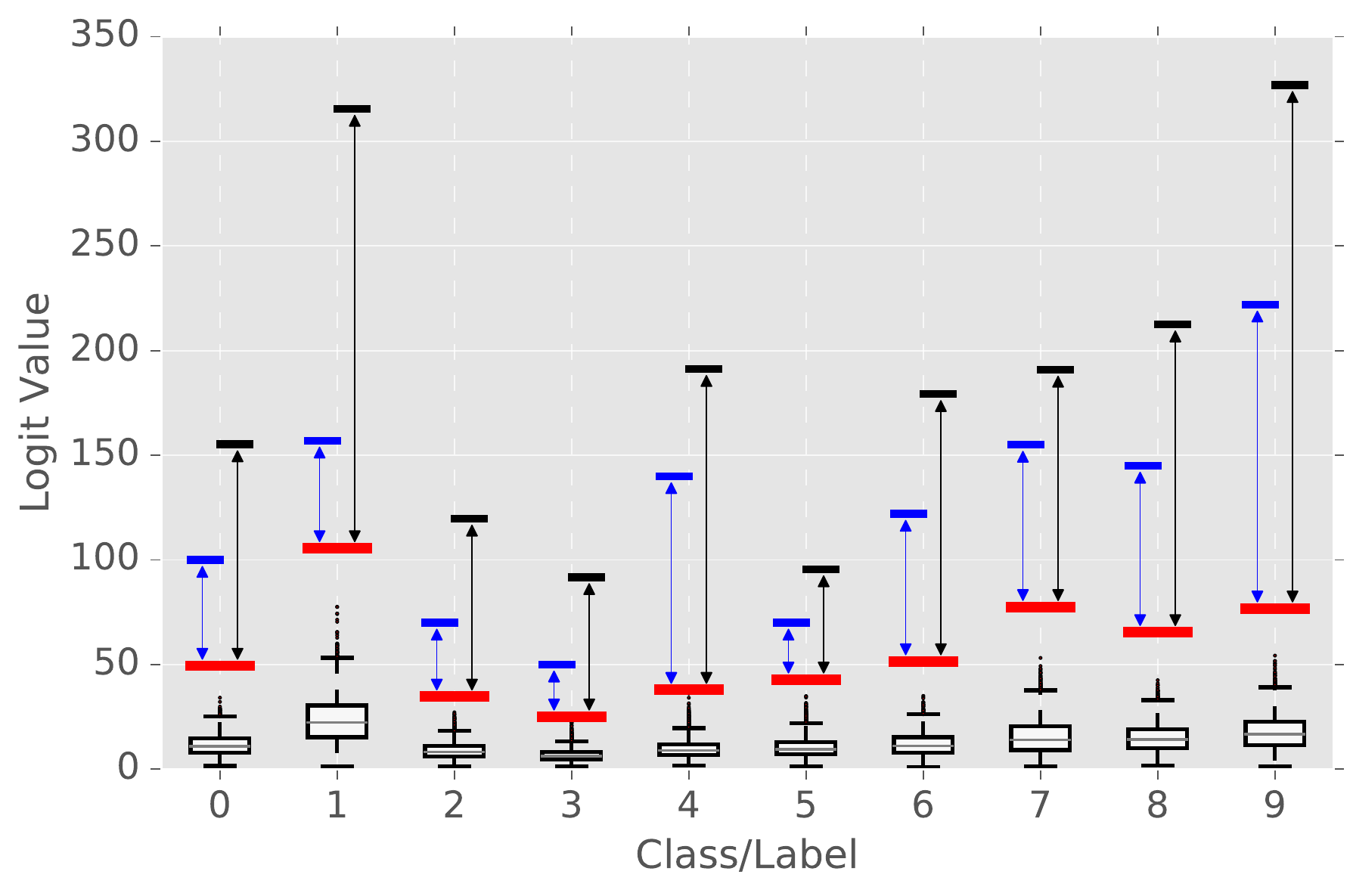}
          \caption{Dataset: CIFAR-10\\Model: Extended LeNet-5}
      \end{subfigure}
      \caption{Illustration of adversarial subspaces with high logit values for (a) MNIST and (b) CIFAR-10. Logit value distributions of the predictions for the genuine images from the training set are given in the form of boxplots. The calculated thresholds and the highest logit value of the adversarial examples generated with I-FGS and the CW are highlighted as red, blue, and black lines, respectively. The adversarial spaces found using our method are indicated with arrows. Best viewed in color.}
      \label{fig:mnist-cifar}
      \vspace{-1em}
\end{figure*}


\subsection{Determining the Threshold}
\label{Determining}
Considering that datasets only contain a limited number of representations (per class), it is highly unlikely that the hypothetical genuine image that attains the highest activation is present in the dataset. However, we can estimate a threshold based on the activations of the observations at hand. A critical point in determining this threshold is to make sure that none of the existing images in the dataset is labeled as adversarial, as it is more important not to cast doubt on good observations than to miss an outlier \cite{doi:10.1080/00031305.1989.10475612, doi:10.1080/00401706.1985.10488027}. The same point is also highlighted by \cite{DBLP:journals/corr/CarliniW17} when evaluating defense mechanisms.


As we showed previously, the logit value distributions are vastly different. Hence, when identifying outliers, we want to avoid any method that (implicitly or explicitly) makes assumptions on the data distribution. Therefore, based on the idea of boxplots, we propose using the interquartile range (IQR) to determine the threshold for identifying outliers (i.e., identifying adversarial examples). The IQR is defined as the difference between the 75th percentile ($Q_3$) and the 25th percentile ($Q_1$): $IQR = Q_3 - Q_1$. In basic statistical analysis, outliers are generally defined as those points that lie beyond the whiskers of the boxplot (i.e., below $Q_1 - k \ IQR$ or above $Q_3 + k \ IQR$). Traditionally, $k$ is set to 1.5 \cite{Tukey-book}, in which case the outliers are referred to as \textit{mild}. We are, of course, only interested in large positive outliers. In this context, the authors of \cite{doi:10.1080/00031305.1989.10475612}, \cite{doi:10.1080/00401706.1985.10488027}, and \cite{Tukey-book} recommend using $k=3$ for determining \textit{extreme} outliers (i.e., highly unusual data points). Since we do not exactly know the underlying distribution of the logits, we experiment with different $k$ values.


For each class, we calculate $Q_3 + k \ IQR$ as the threshold from the training sets of the aforementioned datasets and we present the percentage of misidentified images for the test datasets. Specifically, in Table~\ref{t:k-table}, we show the percentage of genuine images that are misindentified as \textit{outliers} (i.e., adversarial examples) by the calculation $g(\theta , \mathbf{x}) \overset{?}{>} Q_3 + k \ IQR$, for different values of $k$. In the case of ImageNet, we present results for three different architectures, namely, VGG-16, ResNet-50, and Inception-V3 \cite{VGG, resnet, szegedy2015going}. Based on these results, we can observe that $k=3$ is mostly sufficient for most of the architectures, as the percentage of misidentified genuine images is, at most, as low as $0.07\%$ (this corresponds to only $35$ images in the test dataset of ImageNet). Nevertheless, it is also reasonable to prefer a threshold that does not reject \textit{any} genuine image at all. In that regard, we can easily find $k_{\min}$ for different datasets and models, given that the proposed method is non-parametric in nature, which makes it straightforward to adopt this method for different problems. Additionally, by using the IQR instead of parameters such as the mean or standard deviation, which are sensitive to outliers and which make implicit assumptions about the underlying distributions, the proposed method is also more robust (i.e., not attracted by outliers). In the next section, we present experiments that show how the proposed method can be effectively used to detect over-optimized adversarial examples.
\begin{figure}
  \includegraphics[width=0.49\textwidth]{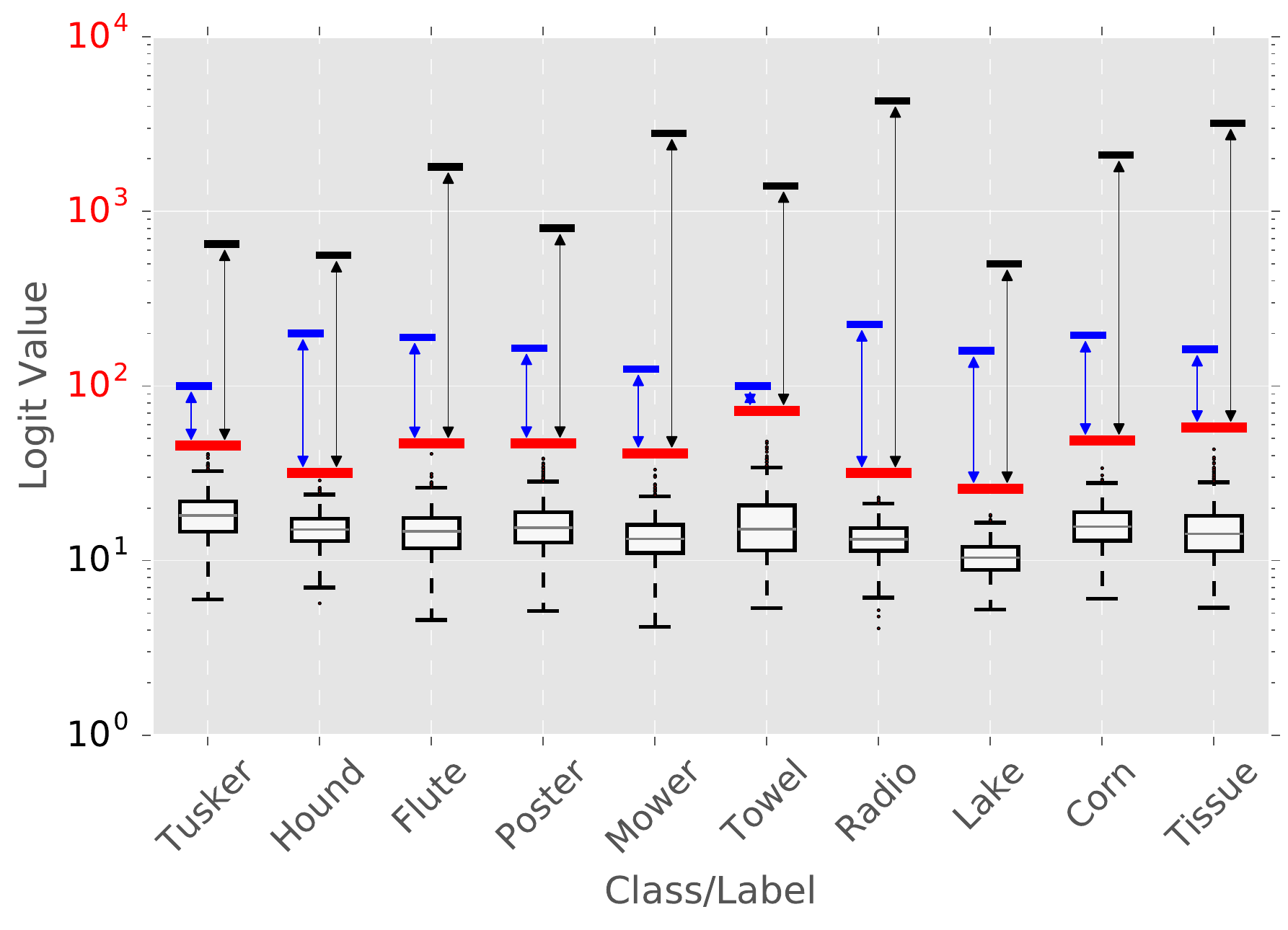}
          \caption{Illustration of adversarial subspaces with high activation for the ImageNet dataset using ResNet-50. Logit value distributions of the predictions for the genuine images from the training set are given in the form of boxplots. The calculated thresholds and the highest logit value of the adversarial examples generated with I-FGS and the CW are highlighted as red, blue, and black lines, respectively. The adversarial spaces found using our method are indicated with arrows. Note that the $y$-axis is log-scaled for clarity. Best viewed in color.}
          \label{fig:imagenet}
          \vspace{-1em}
 \end{figure}
\section{Experimental Results}
\label{Experimental Results}
Using $T = Q_3 + k \ IQR$, we calculated logit thresholds that do not leave out any genuine images for MNIST, CIFAR-10, and ImageNet, using the $k_{\min}$ values given in Table~\ref{t:k-table}. For each of the selected classes in those datasets (all classes in the case of MNIST and CIFAR-10; ten randomly selected classes in the case of ImageNet), we generate $500$ adversarial examples using I-FGS and the $L_2$ version of the CW, totalling up to $5000$ adversarial examples for each dataset, with the aim of finding an adversarial example that generates the highest logit value for that class. By doing so, we want to determine the space for which we can identify \textit{all} adversarial examples that lie between the proposed threshold and the produced logit limit for the adversarial example generation methods.

Fig.~\ref{fig:mnist-cifar} and Fig.~\ref{fig:imagenet} show the results obtained for MNIST, CIFAR-10, and ImageNet (with ResNet-50), with the predicted logit values given in the form of boxplots and with the calculated thresholds highlighted with red lines. The adversarial examples that generate the highest logit prediction values are highlighted separately for the two adversarial example generation methods, using a blue line for I-FGS and a black line for CW. We annotate the space between the proposed threshold (red) and the maximum logit value for each adversarial example generation method $g_{c}(\theta , \mathbf{x}) > T$ to show that we can immediately identify numerous adversarial examples within this space, no matter which method is used to generate these adversarial examples. As expected, for all three datasets, the CW generates \textit{stronger} adversarial examples that produce higher logit values than I-FGS. 


Note that, as the image resolution increases, the upper limit for the logit values that can be produced by an adversarial example also increases. From MNIST to CIFAR-10, this limit increases by a factor of $7$, and from MNIST to ImageNet, this limit increases by a factor of $200$. In the case of the ImageNet dataset, the increase is so high that we present the $y$-axis of the graph in the base ten logscale for the sake of readability.

\subsection{Feasibility for Higher Resolution Images}
\label{Larger Images}


As shown by Fig.~\ref{fig:mnist-cifar} and Fig.~\ref{fig:imagenet}, when the resolution of an image increases, the space in which adversarial examples can be generated also increases in a significant way. This property is also highlighted by other studies \cite{adversarial_input_size}. 

Let us define, for a given model, the effectiveness of our method as
\begin{align}
\label{eq:Adv_calc}
D_{Adv} = \frac{1}{M} \displaystyle\sum_{c=1}^{M} \frac{g(\theta , (\mathbf{x}_{Adv})_c)_{c} - T_c }{g( \theta , (\mathbf{x}_{Adv})_c)_{c}},
\end{align}
where $M$ is the number of classes in the dataset, $c$ is the current target class, $(\mathbf{x}_{Adv})_c$ is the adversarial example that produces the highest logit value when targeting class $c$, and $T_c = (Q_3 + k_{min} \ IQR)_c$ is the calculated threshold for the targeted class. For each class, we calculate the space between the threshold $T_c$ and the logit value of $(\mathbf{x}_{Adv})_c$ and normalize it by dividing by the length of the total space. This value corresponds to the proportion of the adversarial space for class $c$ with respect to the size of the total target class that we can detect by our method. We then take the average over all classes to get an approximate idea of the proportion of the adversarial space we can detect in function of the total dataset. Thanks to the normalization factor, $D_{Adv}$ will allow us to compare different dataset/model combinations, where a higher number indicates a higher effectiveness of detecting adversarial examples.

Applying this formula allows us to construct Table~\ref{t:e-table}, which shows the proportion of potential adversarial examples detected. Based on the results presented in Table~\ref{t:e-table}, we observe that the proposed method is able to detect more adversarial examples when the resolution of an image is higher.


\begin{table}[t!]
  \centering
  \begin{tabular}{lcccccc}
    \toprule
    Dataset (Model)   & Image & $k_{min}$  & Space Covered \\ &Resolution&& $D_{Adv} \in [0, 1]$ \\
    \midrule
    MNIST (LeNet-5) & $1\times28\times28$ & $2.2 $& $26\%$\\
    CIFAR-10  (Ext. LeNet-5)& $3\times32\times32$ & $4.9$ & $68\%$  \\
    ImageNet (VGG-16) &$3\times224\times224$ &  $4.6$ &$75\%$ \\
     ImageNet (ResNet-50) &$3\times224\times224$ &  $3.9$ & $79\%$\\
      ImageNet (Inception-v3)&  $3\times299\times299$ & $4.9$ & $89\%$\\
    \bottomrule
  \end{tabular}
  \vspace{0.3em}
    \caption{Approximate proportion of the number of adversarial examples detected in each dataset (calculated using Eq.~\ref{eq:Adv_calc}).}
\label{t:e-table}
\vspace{-1em}
\end{table}

\subsection{Scalability and Computational Cost}


In Section~\ref{Determining}, we noted that the thresholds must be calculated individually for each class. Naturally, when the number of classes increases, the number of thresholds that must be calculated also increases. However, since these thresholds only need to be calculated and stored once, the computational cost of the proposed method is small; it only requires one full forward pass over the train dataset, after which the thresholds can be used indefinitely. 

Another strength of the proposed method is the easiness with which  thresholds can be updated. When the size of a training set increases (thanks to an influx of additional data), the thresholds can simply be re-calculated.

\subsection{Defense Against Adversarial Examples}
\label{Practicality}


In Section~\ref{Experimental Results} and Section~\ref{Larger Images}, we put the emphasis on the number of potential adversarial examples countered using the proposed method. However, only using the proposed method is not a viable option to identify all adversarial examples. This is especially true in white-box settings, for which it is easy to come up with an attack that would bypass our method. Indeed, one can simply introduce a logit constraint when generating adversarial examples, making sure the generated adversarial examples produce lower logit values than the proposed thresholds. Nonetheless, our intention in proposing this method is not to use it as a universal method for detecting all adversarial examples, but rather to add it to a particular defense workflow as a first (and computationally inexpensive) line of defense for identifying adversarial examples, with these adversarial examples producing unnatural logit values that are beyond the reach of any genuine images. 

Applying the proposed method ensures that adversarial example generation methods will be limited not only by the amount of perturbation and the discretization constraint, but also by the maximum logit value produced by an adversarial example. This will limit the search space extensively, especially when the resolution of the images under consideration is large, as is shown in Section~\ref{Larger Images}.

\section{Conclusion and Future Research}

In this paper, we established the fundamentals for the use of logit values as an indicator of adversariality. To that end, we first discussed the masking effect of the softmax function, showing that the logit values keep increasing, even after softmax output achieving its maximum value of one during the generation of adversarial examples.

Next, we laid out examples of logit distributions from multiple datasets and showed the need for a distribution-free method to identify adversarial examples, leading to the introduction of a non-parametric and computationally cheap technique for detecting over-optimized adversarial examples. Throughout this paper, we presented experimental results for MNIST, CIFAR-10, and ImageNet, also using different neural network architectures.

The main purpose of the newly introduced defense technique is to further limit the expressiveness of methods for adversarial example generation, not only in terms of perturbation amount, but also in terms of logits. This allows our technique to be used as a first line of defense, before triggering a well-rounded one that is more complex in nature. Therefore, future research may focus on investigating the compatibility of our technique with other defense mechanisms that are complimentary in nature.

\vspace{-0.34em}
\section*{Acknowledgements}

The research activities described in this paper were funded by Ghent University Global Campus, Ghent University, imec, Flanders Innovation \& Entrepreneurship (VLAIO), the Fund for Scientific Research-Flanders (FWO-Flanders), and the EU.
\bibliographystyle{IEEEtran}
\bibliography{2018_10}

\end{document}